\documentclass[conference]{IEEEtran}
\IEEEoverridecommandlockouts
\usepackage{cite}
\usepackage{amsmath,amssymb,amsfonts}
\usepackage{algorithmic}
\usepackage{graphicx}
\usepackage{textcomp}
\usepackage{xcolor}
\def\BibTeX{{\rm B\kern-.05em{\sc i\kern-.025em b}\kern-.08em
    T\kern-.1667em\lower.7ex\hbox{E}\kern-.125emX}}
\begin{document}

\title{Demo Abstract: Real-Time Out-of-Distribution Detection on a Mobile Robot
\thanks{This research was funded in part by MoE, Singapore, Tier-2
grant number MOE2019-T2-2-040.  This research is part of the programme DesCartes and is supported by the National Research Foundation, Prime Minister’s Office, Singapore under its Campus for Research Excellence and Technological Enterprise (CREATE) programme.}
}

\author{\IEEEauthorblockN{Michael Yuhas\textsuperscript{1,2}, Arvind Easwaran\textsuperscript{1}}
\IEEEauthorblockA{\textit{\textsuperscript{1}School of Computer Science and Engineering} \\
\textit{\textsuperscript{2}Energy Research Institute @ NTU, Interdisciplinary Graduate Program}}
\textit{Nanyang Technological University}, Singapore \\
michaelj004@e.ntu.edu.sg, arvinde@ntu.edu.sg
}

\maketitle

\begin{abstract}
In a cyber-physical system such as an autonomous vehicle (AV), machine learning (ML) models can be used to navigate and identify objects that may interfere with the vehicle's operation.  However, ML models are unlikely to make accurate decisions when presented with data outside their training distribution.  Out-of-distribution (OOD) detection can act as a safety monitor for ML models by identifying such samples at run time. However, in safety critical systems like AVs, OOD detection needs to satisfy real-time constraints in addition to functional requirements.  In this demonstration, we use a mobile robot as a surrogate for an AV and use an OOD detector to identify potentially hazardous samples. The robot navigates a miniature town using image data and a YOLO object detection network.  We show that our OOD detector is capable of identifying OOD images in real-time on an embedded platform concurrently performing object detection and lane following.  We also show that it can be used to successfully stop the vehicle in the presence of unknown, novel samples.
\end{abstract}

\section{Introduction}
\begin{figure}[htbp]
    \centering
    \includegraphics[width=0.485\textwidth]{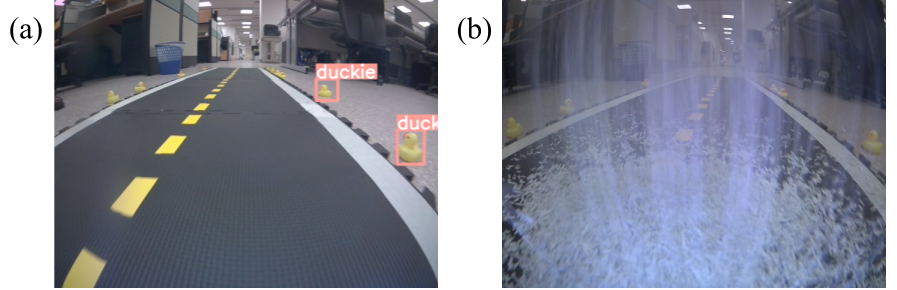}
    \caption{(a) The YOLO (you only look once) model detects the two nearest ducks when an image is ID; (b) the YOLO model is unable to identify any objects in the OOD image with simulated snowfall.}
    \label{fig:id_ood}
\end{figure}

Machine learning (ML) models are not likely to perform well when they receive samples outside of their training data distributions.  This poses a major risk to safety critical cyber-physical systems such as autonomous vehicles (AVs).  Consider an ML-based object detector deployed to an AV: during training, the object detector may be exposed to images with little or no snow, and during operation, a heavy snowstorm could lead to poor results.  Fig.~\ref{fig:id_ood}-a shows an in-distribution (ID) image for such a system: the YOLOv7 model~\cite{wang2022yolov7} successfully detects the ducks closest to the vehicle when no snow is present.  However, when excessive snowfall occurs (simulated by falling confetti) the scene becomes out-of-distribution (OOD) (Fig.~\ref{fig:id_ood}-b), and the model can no longer detect the ducks.  It is imperative that such OOD samples are identified to prevent the system from taking dangerous control actions.  An OOD detector can be used as a run-time safety monitor to achieve this goal, however, the detector must also meet hard real-time deadlines to ensure that detection occurs with sufficient time to avoid a collision~\cite{yuhas2021embedded}.  Furthermore, OOD detection on image data relies on deep neural networks, which must share computational resources with other safety critical tasks.

Prior works have deployed OOD detectors to mobile robots and demonstrated their ability to meet deadlines and detect harmful samples at run time.  In~\cite{cai2020realtime}, the reconstruction loss of a variational autoencoder (VAE) was used to perform OOD detection on image data in a simulated environment.  Both detection accuracy and execution time were considered, however, the experiments were not performed on a physical system.  In~\cite{yuhas2021embedded}, the latent space of a VAE was used for OOD detection on a mobile robot to trigger an emergency stop before a collision occurred.  While this work was demonstrated on a physical system, the OOD detector failed to meet real-time constraints and was not always able to stop the robot in time.  In~\cite{burruss2021deeprbf}, a deep radial basis function (RBF) network was trained to steer a mobile robot around a racetrack and simultaneously perform OOD detection.  This integrated OOD detection and steering control network requires regression tasks to be reformulated as classification tasks, which is not always desirable.  So far, no work has deployed a deep ML model and an independent OOD detector to a CPS and observed the effect on the response times of both tasks.

We will perform a live demonstration of the feasibility of an OOD detector as a real-time safety monitor for a mobile robot performing lane navigation and object detection tasks.

\begin{enumerate}
    \item Our OOD detector meets real-time deadlines \textit{and} does not interfere with the lane follower's operation.
    \item Our OOD detector successfully triggers emergency braking in OOD conditions without excessive false-positives.
    \item Our test bed allows us to easily create unique test scenarios; while we consider the case of snowfall as OOD, the audience can attempt to foil the OOD detector with other conditions.
\end{enumerate}

\section{System Design}
\begin{figure}[htbp]
    \centering
    \includegraphics[width=0.48\textwidth]{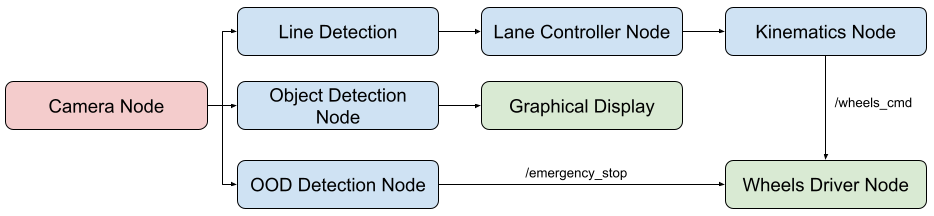}
    \caption{Software system block diagram of our Duckiebot.  Red indicates a sensing node, blue corresponds to a signal processing node, and green represents an output node.}
    \label{fig:sysbd}
\end{figure}

We use the Duckietown platform~\cite{liam2017duckietown} to simulate an AV.  Duckietown consists of small, holonomic robots (Duckiebots) that navigate a miniature world (Duckietown) with lane markings that mimic actual roads.  The inhabitant of Duckietown (rubber ducks) serve as obstacles that the Duckiebot must not hit.  This platform allows us to deploy multiple ROS (Robot Operating System) packages to a Duckiebot, each running within its own Docker container. Fig.~\ref{fig:sysbd} shows the high-level software block diagram of our system. A lane follower based on traditional computer vision (CV) techniques identifies road markings, calculates the desired pose of the robot, and sends this information to a kinematics node that calculates the wheel rotations required to achieve that pose. In parallel, an object detection node identifies potential obstacles and sends bounding box coordinates to a graphical display for visualization.  An OOD detector acts as a safety monitor for the entire system and sends an emergency stop message if an incoming image is OOD.  The system has one sensor (camera node) and two actuators (wheels driver node).  The source code for our software implementation is available on GitHub. \footnote{https://github.com/CPS-research-group/CPS-NTU-Public}

\subsection{Lane Following}
Duckietown's native lane following package is used to steer the robot~\cite{liam2017duckietown}.  It uses traditional CV algorithms and consists of the following steps: image equalization, road marking detection, projection from image space to the world frame, a Bayesian filter for lane localization, and finally the lane controller which generates the wheel commands.  In isolation we measured an average-case execution time (ACET) of 40.1~ms and a worst-case execution time of 134.3~ms.  By using traditional CV line detection, we are able to show that the OOD detector is also useful at protecting non-ML components from anomalous images.  For example, under conditions such as heavy snow, lane markings may not be visible and the lane follower will mistake spurious lines in the image as road markings leading to inappropriate control actions.

\subsection{Object Detection}
\begin{figure}[htbp]
    \centering
    \includegraphics[width=0.48\textwidth]{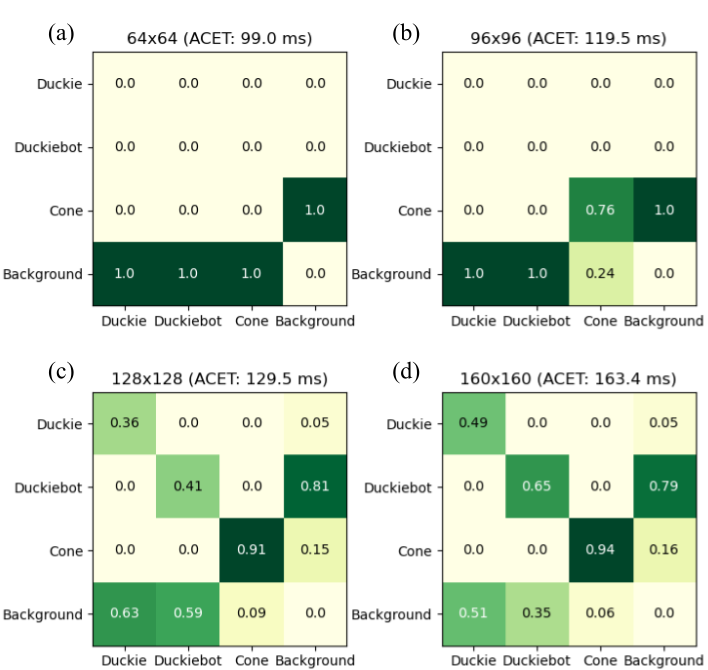}
    \caption{Confusion matrices and corresponding ACETs for a YOLOv7 tiny network trained on the Duckietown dataset with input sizes (a)~64x64, (b)~96x96, (c)~128x128, and (d)~160x160.}
    \label{fig:yolo}
\end{figure}

To accomplish object detection, we trained a YOLOv7 tiny object detection network~\cite{wang2022yolov7} on the Duckietown object detection dataset that contains three classes: rubber ducks, traffic cones, and Duckiebots. YOLOv7 tiny is a lightweight variant of the YOLO family of object detection networks that identifies bounding box coordinates and object class information simultaneously~\cite{redmon2016you}.  Although not as accurate as two-stage methods like Fast R-CNN, YOLO networks are capable of faster inference times, which is desirable in our application. YOLOv7 tiny can be trained for different input image sizes and Fig.~\ref{fig:yolo} shows the resulting trade-off between ACET and the model's ability to reliably identify objects.  All ACETs are measured with the model quantized to 8-bits via dynamic quantization~\cite{krishnamoorthi2020introduction} and the QNNPACK backend for inference on the AArch64 target platform~\cite{dukhan2019}. The YOLOv7 tiny model trained on 64x64 images (Fig.~\ref{fig:yolo}-a) is unable to make any correct detections, but has the fastest ACET.  We select the model trained on 160x160 images (Fig.~\ref{fig:yolo}-d) for use in our demo: it is better at identifying the rubber ducks, but has an ACET of 163.4~ms.

\subsection{Out-of-Distribution Detection}
In order to construct an accurate, yet fast OOD detector, we used the OOD design methodology proposed in~\cite{yuhas2022design}.  This methodology takes an existing OOD detector and searches preprocessing parameter combinations and quantization schemes to find several candidate solutions.  We plan to focus on OOD caused by falling snow (Fig.~\ref{fig:id_ood}-b), so we started with the optical flow OOD detector proposed in~\cite{feng2021improving} that uses a variational autoencoder to identify environmental motion not present in the training set.  Our ID dataset consisted of 2048 images gathered by the Duckiebot navigating an empty track autonomously.  We converted sequential images into optical flow matrices using the Farneb{\"a}ck optical flow algorithm and ran a genetic algorithm (GA) to select the input image size, input depth (sequential flows), and interpolation method that maximize the F1 score of the OOD detector.  To calculate F1 score, we used a test set containing eight videos with frames labeled ID or OOD based on the presence of simulated snowfall. Table~\ref{table:sol} shows the candidate solutions identified by the GA and their respective functional and non-functional performance.  All the execution times (ETs) are calculated with 8-bit dynamic quantization~\cite{krishnamoorthi2020introduction} and tested with the QNNPACK backend~\cite{dukhan2019} on the target platform.  We selected the best OOD detector identified by the GA (60x80 input size, 10 flows, bilinear interpolation) for our demo.

\begin{table}[htbp]
    \centering
    \caption{Candidate solutions for the OOD detector with their functional performance (F1 score) and ET mean and variance.} \label{table:sol}
    \begin{tabular}{|c|c|c|c|c|c|}  
    \hline
    \textbf{Size} & \textbf{Flows} & \textbf{Interp.}  & \textbf{F1 Score} & \textbf{ET (mean)} & \textbf{ET (var.)}\\
    \hline
    30x40 & 4 & Bilinear & 0.43 & 44.1 ms & 1098 ms$^2$\\
    \hline
    60x80 & 5 & Bilinear & 0.97 & 53.3 ms & 1485 ms$^2$\\
    \hline
    90x120 & 2 & Bilinear & 0.54 & 58.1 ms & 900 ms$^2$\\
    \hline
   120x160 & 12 & Bilinear & 0.90 & 70.0 ms & 1096 ms$^2$\\
   \hline
    
\end{tabular}
\end{table}

\section{Demonstration}
\begin{figure}[htbp]
    \centering
    \includegraphics[width=0.48\textwidth]{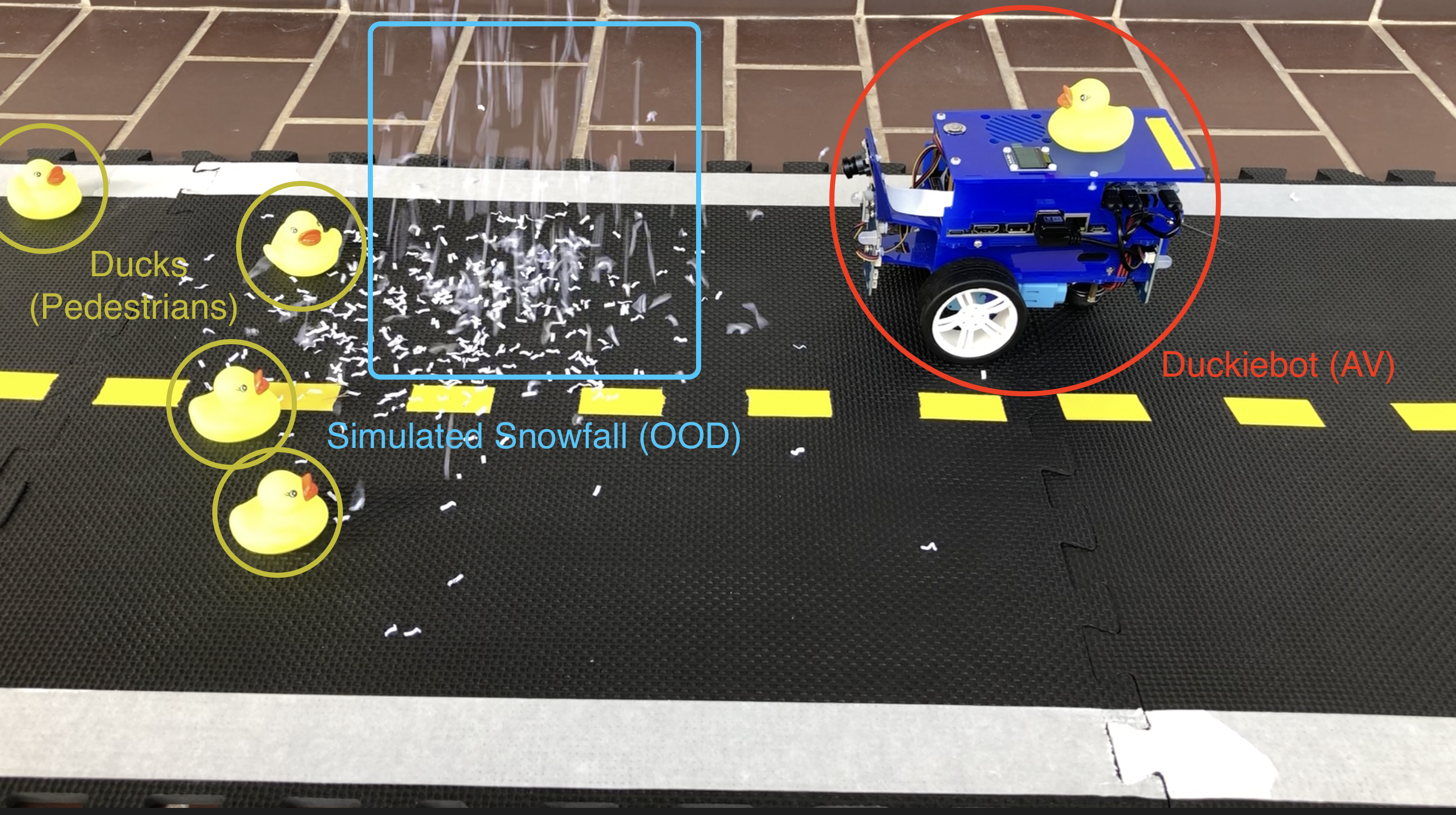}
    \caption{The main components of the experimental setup: the Duckiebot (red) is a surrogate for an AV, the ducks (yellow) simulate pedestrians, and the confetti (blue) simulates snowfall.}
    \label{fig:exp}
\end{figure}

\begin{figure}[htbp]
    \centering
    \includegraphics[width=0.48\textwidth]{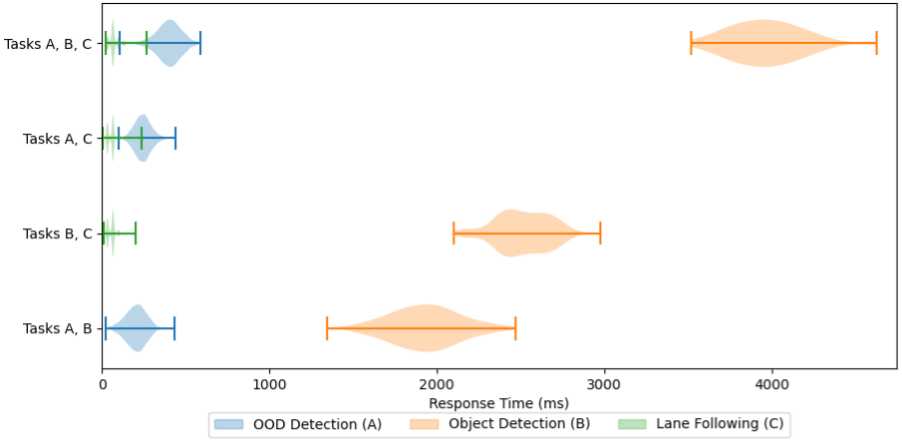}
    \caption{Response time distributions when the OOD detection task (A), object detection task (B), and lane following task (C) run simultaneously on the Duckiebot.  The response times of all three tasks increase in comparison to their ETs, but object detection suffers the most degradation.}
    \label{fig:experiment}
\end{figure}

All experiments were performed on a DB21M Duckiebot equipped with a Jetson Nano 2GB running L4T 32.1 with the PREEMPT\_RT kernel patch installed.  Fig.~\ref{fig:exp} shows the main components of our demonstration.  A Duckiebot drives forward along a road while the object detection network identifies duckies in the environment.  To simulate snowfall, confetti is dumped in front of the robot.  During the demonstration:
\begin{enumerate}
    \item The OOD detector successfully triggers an emergency stop when snowfall is present in the environment.
    \item In the absence of OOD samples, the vehicle can reach the end of the track without a false positive detection.
    \item The OOD detector maintains a response time of less than 800~ms and the lane follower still meets its deadlines.
\end{enumerate}
Fig.~\ref{fig:experiment} shows the response times for the three tasks (OOD detection, object detection, and lane following) when run together on the Duckiebot.  We observe that the OOD detector always has a response less than 800~ms and is able to stop the vehicle before a collision or leaving the road.  Furthermore,  the OOD detector does not interfere with the lane follower's response time and the Duckiebot is still able to navigate.  However, the response time of the object detector increases drastically in comparison to its ACET when it shares the CPU with other tasks.  We believe this is due to the system utilization approaching 100 percent and that when all three tasks are run simultaneously, all 2GB RAM have been utilized and memory intensive tasks (like YOLO inference) must continually swap pages in and out of memory as they are scheduled.

\end{document}